\documentclass[]{interact}

\usepackage{fix-cm}
\usepackage{epstopdf}
\usepackage{subcaption}
\usepackage{url}
\usepackage{xcolor}
\usepackage{hyperref}
\usepackage{amsmath}
\usepackage{threeparttable}
\usepackage{optidef}
\usepackage{algorithmicx}
\usepackage[ruled]{algorithm}
\usepackage[noend]{algpseudocode}

\usepackage{amssymb}
\usepackage{pifont}

\newcommand{\figlab}[1]{\label{fig:#1}}
\newcommand{\figref}[1]{Fig.~\ref{fig:#1}} 

\usepackage[numbers,sort&compress]{natbib}
\bibpunct[, ]{[}{]}{,}{n}{,}{,}
\renewcommand\bibfont{\fontsize{10}{12}\selectfont}
\makeatletter
\def\NAT@def@citea{\def\@citea{\NAT@separator}}
\makeatother

\theoremstyle{plain}

\theoremstyle{definition}

\theoremstyle{remark}

\begin{document}

\articletype{FULL PAPER}

\title{Industrial-Grade Robust Robot Vision for Screw Detection and Removal under Uneven Conditions}

\author{
\name{Tomoki Ishikura\textsuperscript{*a,b}\thanks{\textsuperscript{*}Corresponding Author: Tomoki Ishikura. Email: ishikura.tomoki@jp.panasonic.com}, Genichiro Matsuda\textsuperscript{a}, Takuya Kiyokawa\textsuperscript{b} and Kensuke Harada\textsuperscript{b,c}}
\affil{\textsuperscript{a}Manufacturing Innovation Division, Panasonic Holdings Corporation, 2-7 Matsuba-cho, Kadoma, Osaka, Japan; \\\textsuperscript{b}Department of Systems Innovation, Graduate School of Engineering Science, The University of Osaka, 1-3 Machikaneyama, Toyonaka, Osaka, Japan; \\\textsuperscript{c}Industrial Cyber-physical Systems Research Center, The National Institute of Advanced Industrial Science and Technology (AIST), 2-3-26 Aomi, Koto-ku, Tokyo, Japan.}
}

\maketitle

\begin{abstract}
As the amount of used home appliances is expected to increase despite the decreasing labor force in Japan, there is a need to automate disassembling processes at recycling plants. 
The automation of disassembling air conditioner outdoor units, however, remains a challenge due to unit size variations and exposure to dirt and rust. 
To address these challenges, this study proposes an automated system that integrates a task-specific two-stage detection method and a lattice-based local calibration strategy. 
This approach achieved a screw detection recall of 99.8\% despite severe degradation and ensured a manipulation accuracy of $\pm$0.75 mm without pre-programmed coordinates. 
In real-world validation with 120 units, the system attained a disassembly success rate of 78.3\% and an average cycle time of 193 seconds, confirming its feasibility for industrial application.
\end{abstract}


\section{Introduction}
The global discharge of waste electrical and electronic equipment (WEEE) has reached approximately 50 million tons per year due to mass-production and mass-disposal-oriented economic activities, leading to environmental problems such as resource depletion~\cite{Balde2022}.
To address this issue and promote the establishment of a recycling-oriented society, the Home Appliance Recycling Law was enforced in Japan in April 2001~\cite{MATSUTO2004425}. This law mandates manufacturers to recycle used home appliances.
Among recycling processes, horizontal recycling, which refers to high-grade resource circulation reusing materials from used home appliances for new products, is particularly important for realizing a sustainable society.

Notably, the number of air conditioners processed reached 3.69 million units in fiscal year 2022~\cite{aeha2022report}, driving an urgent need for robotic automation to alleviate labor shortages and operational risks. 
Achieving automation is challenging because fasteners, such as screws and bolts, often account for 30\% to 50\% of the total component count in WEEE~\cite{Boothroyd2010}.
Consequently, the removal of these fasteners constitutes the primary bottleneck in the disassembly process.
In this context, conventional single-stage methods face inherent limitations in generalizing across the high visual variability of WEEE fasteners. Severe degradation, including rust and physical deformation, results in inconsistent feature representations, leading these models to frequently misclassify background surface irregularities that share similar visual features with the degraded fasteners as target components.

This study designs and implements an automated system for the reversible disassembly of air conditioner outdoor units, targeting the removal of fastening components and exterior housing in the upstream process. 
To achieve industrial viability, we address three critical challenges: (1) achieving high-recall detection across a wide variety of diverse, degraded product models without relying on time-consuming sequential imaging protocols, (2) ensuring manipulation accuracy within $\pm$0.75 mm for distorted workpieces without pre-programmed coordinates, and (3) meeting a cycle time of $\leq 216$ seconds with a completion rate of $> 75\%$, as defined in the problem setting (Section 3.).
The contributions of this work are threefold:
(1) We propose a task-specific two-stage detection method robust against variable visual conditions, which coarsely detects candidate regions with visual similarities to fasteners in the first stage and precisely extracts geometric features in the second stage.
(2) We propose a robust calibration technique to mitigate spatially varying, nonlinear error sources, specifically residual lens distortion and robot structural deflection.
(3) We demonstrate the system's practical feasibility through extensive real-world experiments involving the disassembly of 120 air conditioner outdoor units sourced from an actual recycling plant. 

\section{Related works}
\subsection{Circular economy and material recovery}
WEEE is a major cause of resource depletion and environmental pollution, and advanced material recovery technologies are required from the perspective of the circular economy~\cite{Balde2024GEM,electronics11040533}. Previous studies have reported that high-purity recovery of plastics, metals, and rare metals, as well as the precise extraction of reusable components, contributes to increasing the value of secondary materials and reducing CO$_2$. emissions~\cite{RAVI2012145,HONG2015357,MENIKPURA2014183}. 
In addition, since WEEE includes a wide variety of product types and model variations, recycling plants require flexible disassembly technologies capable of handling diverse products~\cite{ASIF2024483}. 
WEEE often suffers from severe physical degradation, including corrosion, structural deformation, and surface contamination, which introduce significant uncertainty into perception and manipulation tasks.
These uncertainties require cognitive capabilities to dynamically assess the product state~\cite{Vongbunyong2015}. 
Also, disassembly involves contact-rich manipulation, such as unscrewing and unfastening snap-fits, which demands sub-millimeter precision despite positional ambiguities caused by deformation~\cite{Zhang2022}. 
Therefore, the robot adapts to the physical condition of each fastener, making robust detection and calibration critical.

Studies on the taxonomy of fasteners in WEEE reveal that screws and snap-fits are the predominant joint types encountered across various appliance categories~\cite{Boothroyd2010,Li2020}. 
Furthermore, time analyses of manual disassembly processes consistently identify the identification and removal of these fasteners as the most time-consuming operations, accounting for a significant portion of the total cycle time~\cite{Vanegas2018}. 
Therefore, developing a robust robotic solution for fastener removal is not merely a technical improvement but a critical requirement to reduce operational costs and meet industrial throughput targets.

For this reason, robotic automation that enables precise and high-quality disassembly and is capable of handling diverse products is attracting attention as a key technology to improve both the quality and quantity of recovered materials.

\subsection{Elemental technologies for screw unfastening}
Automated disassembly robots for WEEE have been investigated in several studies. DiFilippo et al.~\cite{app14146301} proposed a single-stage neural network-based detection method for laptop computers used in clean environments, achieving a 95\% detection rate.
The method is promising but requires further robustness against rusted or contaminated screws common in outdoor-installed equipment. Similarly, other deep-learning approaches~\cite{Li2020,Zhang2022,Zhang2023,Ueda2024,Deng2024,Diaz2025} have largely overlooked detection performance under degraded conditions (e.g., dirt, rust, deformation). 

Some end-effectors have been designed to compensate for detection errors~\cite{Zhang2023,al2024automated,robotics11010018}. These end-effectors incorporate physical compensation mechanisms to mitigate localization and control errors. For instance, the integration of elastic components, such as springs or rubber elements, enables the screwdriver bit to passively align with the screw head upon contact, facilitating a self-centering effect. Additionally, the implementation of conical guides around the bit can physically funnel the screw head into the correct position. By adopting these passive compliance strategies, these designs aim to enhance the system's tolerance for positional errors, thereby seeking to increase the robustness of the robotic process against sensing and motion inaccuracies.

Clark et al.~\cite{app15020618} developed a robotic disassembly system using a single-stage DCNN for detecting cross-recessed screws on a UR5 platform. While the system demonstrates high reliability with a 100\% extraction rate within two attempts and a localization accuracy of $\pm 1$ mm, its throughput is limited by a cycle time spanning 30 to 55 seconds per screw. This significant latency arises from a two-level imaging protocol that necessitates a high-level overview followed by a close-up for accurate localization. Additionally, the process is prolonged by sequential force and current sensing routines required to confirm tool-to-screw engagement. In the context of large-scale industrial appliances, the narrow field of view at short camera ranges forces the robot into multiple repositioning steps. These combined delays in sensing and movement ultimately reduce throughput, presenting a barrier to high-speed industrial implementation.
Also, DiFilippo et al. proposed a multi-modal approach using visual and force feedback; however, it remains insufficient in terms of throughput and robustness for practical industrial disassembly~\cite{Difilippo2018}.

Disassembling the outer housing of an air conditioner outdoor unit requires removing over 20 screws. Conventional methods~\cite{app15020618} would take over 10~min per unit, which is operationally impractical. 
Furthermore, a 95\% per-screw detection rate yields a batch success probability of only $0.95^{20} \approx 36\%$, meaning the process is likely to fail. 
To address these issues, there is a critical need for a system capable of rapid, accurate localization of multiple screws such as weathered ones from a single image, enabling high-throughput batch removal.

\section{Problem setting}
The automated disassembly of air conditioner outdoor units is significantly hindered by the vast diversity of product models and unpredictable physical degradation, including corrosion and deformation.
Consequently, robust detection of such heterogeneous targets remains a formidable challenge for conventional template matching or color-based feature extraction. 
Furthermore, unlike standard industrial automation that relies on high positional repeatability through pre-taught trajectories, disassembly cannot depend on fixed coordinates due to the structural distortion of target objects. Nevertheless, the process demands near zero-miss screw detection to ensure operational reliability.

To address these challenges, we developed a hand–eye robotic system building upon the framework proposed in previous work~\cite{app15020618,Difilippo2018}. 
This system is capable of accurate manipulation through deep learning-based detection and the coordinated use of a wide-view camera and an end-effector-mounted detail camera.
This system required a sequential combination of wide-view imaging (for rough localization) and detailed imaging (for precise positioning). This necessitated multiple cycles of imaging and robot motion, resulting in a processing time of approximately 15 seconds per screw.
Consequently, the disassembly of a single unit, which requires removing an average of 20 screws, took more than 300 seconds. Furthermore, the system suffered from detection errors, including both false positives (misidentifying non-screw features) and false negatives (overlooking actual screws). These inaccuracies not only extended the total processing time but also suggest potential safety risks, posing significant challenges for integration into industrial factory operations.

In practical operations at recycling plants, both cycle time and the disassembly completion rate (the ratio of successfully disassembled units to total processed units) were identified as critical issues. 
The performance targets for the proposed system were established based on practical requirements for industrial recycling plants. According to the FY2024 operational data from the Panasonic Eco Technology Center (PETECK), the annual processing volume for air conditioners is approximately 142,000 units~\cite{peteck2024performance}.
Assuming 240 annual operating days (excluding weekends and holidays), the required daily throughput is approximately 600 units. 
To minimize the facility footprint and optimize capital investment, we selected a five-unit parallel configuration, resulting in a target throughput of 120 units per 8-hour shift per unit. Accounting for a 90\% equipment utilization rate, the required cycle time was determined to be 216 seconds.
Additionally, we aimed to reduce the labor requirement for the 600-unit/day process from 2.0 to 0.5 operators. This implies a target where a single operator oversees multiple units (e.g., 0.1 operators per unit). To achieve this, the disassembly completion rate must exceed 75\% to ensure that the manual intervention for non-automated tasks remains within the capacity of the reduced personnel. Consequently, the primary objectives were defined as a cycle time of $\le 216$ seconds and a completion rate of $> 75\%$.

Achieving these targets required determining how to eliminate the time-consuming multiple imaging steps and improve detection accuracy to minimize false positives.
We targeted a 90\% screw detection rate, a 90\% screw removal success rate, and a 95\% cover removal success rate to ensure the desired disassembly completion rate.

To solve these issues, we propose a robotic screw removal system, presented in Section 4.1. This system utilizes a single wide-view RGB-D camera to capture the entire workpiece. 
Unlike conventional methods, it localizes all screws from a single image acquisition, enabling the robot to remove multiple screws continuously. 
To realize this system, we developed two core technologies: (1) deep learning-based screw detection (Section 4.2), which achieves robustness against degradation such as dirt, rust, and deformation through a task-specific two-stage detection approach, and (2) high-precision robotic manipulation (Section 4.3), which utilizes advanced calibration to compensate for lens distortion, depth measurement errors, and robot arm deflection.
Finally, the practical effectiveness of the proposed system is validated in Section 5 through real-world experimental evaluations, which involve extensive tests using 120 air conditioner outdoor units collected from an actual recycling plant.

\section{Methods}
\subsection{Robotic screw removal system}
\begin{figure}[tb]
    \centering
    \includegraphics[width=0.7\linewidth]{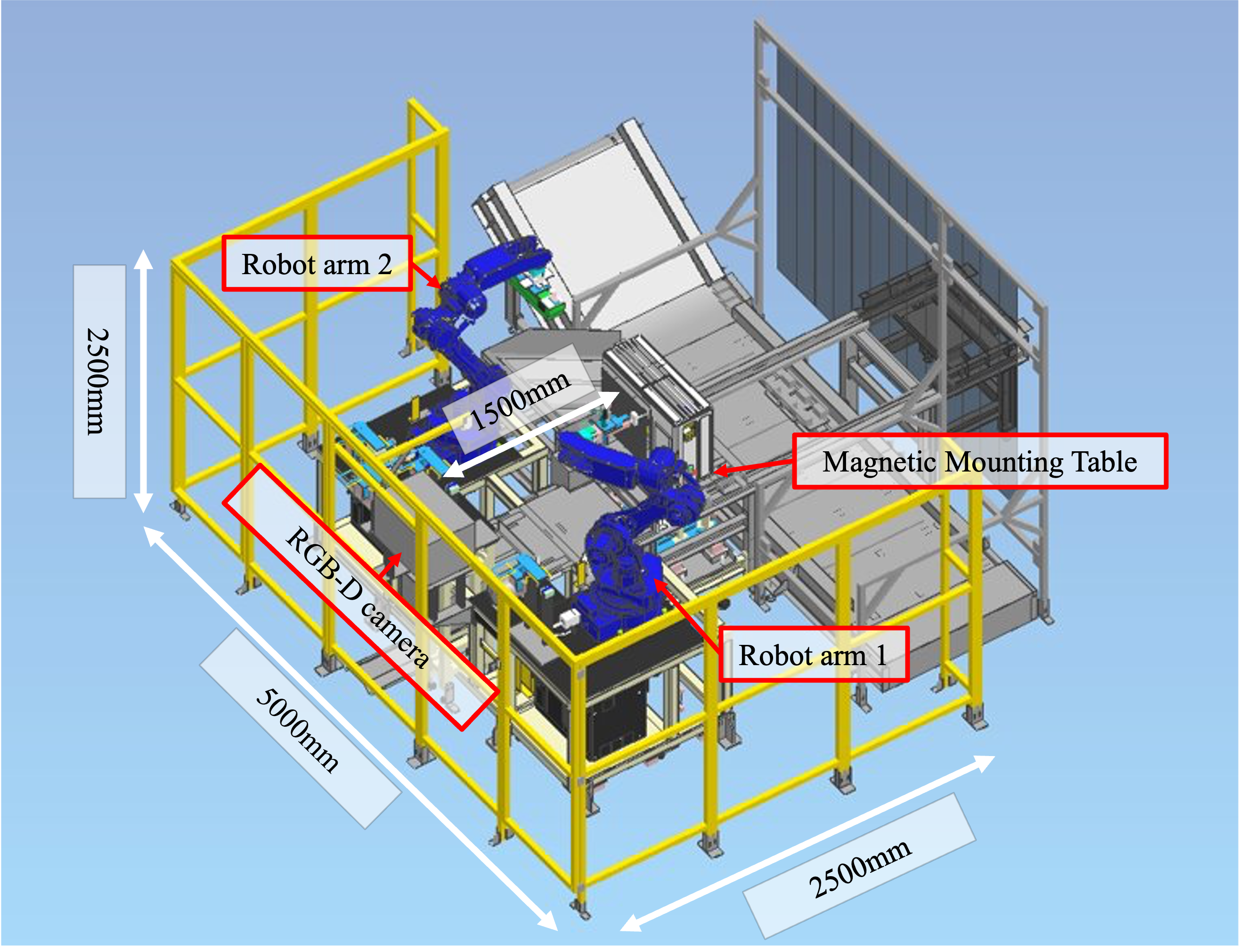}
    \caption{Overview of the automated disassembly system.}
    \figlab{system}
\end{figure}
To reduce cycle time, we consolidated the wide-view and detailed imaging functions, which were previously performed separately, into a single process.
In the proposed configuration, the system performs RGB-image-based screw detection followed by depth-image-based 3D localization, enabling precise positioning across the entire workpiece using wide-view imaging.\figref{system} presents an external view of the developed automated disassembly system.

The system comprises two six-axis articulated robot arms (MOTOMAN-GP25, Yaskawa Electric Corporation)\footnote{MOTOMAN-GP25, Yaskawa Electric Corporation, \url{https://www.e-mechatronics.com/product/robot/assembly/lineup/gp25/spec.html}}, a high-precision wide-view RGB-D camera system, a magnetic mounting table for securing the air conditioner outdoor unit, and a model number reading device. The RGB-D camera system consists of a color CMOS camera (UI-5200SE-C-HQ Rev.4, IDS Imaging Development Systems GmbH)\footnote{UI-5200SE-C-HQ Rev.4, IDS Imaging Development Systems GmbH, \url{https://jp.ids-imaging.com/store/ui-5200se-rev-4-2.html}} with a $4104 \times 3006$ pixel resolution equipped with a 16\,mm focal length lens (LM16FC24M, Kowa Optronics Co., Ltd.)\footnote{LM16FC24M, Kowa Optronics Co., Ltd., \url{https://jp.ids-imaging.com/store/lens-kowa-lm16fc-16-mm-1-1.html}}, paired with a stereo depth camera (Ensenso X36-5FA, IDS Imaging Development Systems GmbH)\footnote{Ensenso X36-5FA, IDS Imaging Development Systems GmbH, \url{https://jp.ids-imaging.com/ensenso-3d-camera-x-series.html}} featuring a $2448 \times 2048$ resolution and a depth accuracy of 0.281\,mm at a distance of 1.5\,m.
The system footprint measures $5000 \times 2500 \times 2500$\,mm (width $\times$ depth $\times$ height), with a working distance of approximately 1500\,mm between the wide-view RGB-D camera and the target unit.
This configuration necessitates only one image acquisition per side, effectively eliminating the robot motion previously required for close-up detailed imaging.
The system can remove 4 to 20 screws from one side within 15-40 seconds, achieving a processing throughput 4 to 7.5 times faster than our conventional system.

\figref{proposal} illustrates the system's workspace and \figref{challenges} illustrates the associated challenges.
Achieving accurate detection and manipulation using solely wide-view imaging requires a total system accuracy of $\pm 0.75$\,mm at any location within the extensive workspace measuring $900 \times 400 \times 750$\,mm (width $\times$ depth $\times$ height). 
To meet this $\pm 0.75$\,mm requirement, the screw detection algorithm must detect screws under varying conditions, such as dirt, rust, or deformation, and estimate the cross-slot center within $\pm 2$ pixels ($\pm 0.4$\,mm) using only a $34 \times 34$ pixel resolution.
The robotic manipulation system requires rigorous calibration to compensate for lens distortion, depth measurement errors, and robot arm deflection, ensuring a mechanical positioning accuracy within $\pm 0.35$\,mm.

\begin{figure}[tb]
    \centering
    \includegraphics[width=0.8\linewidth]{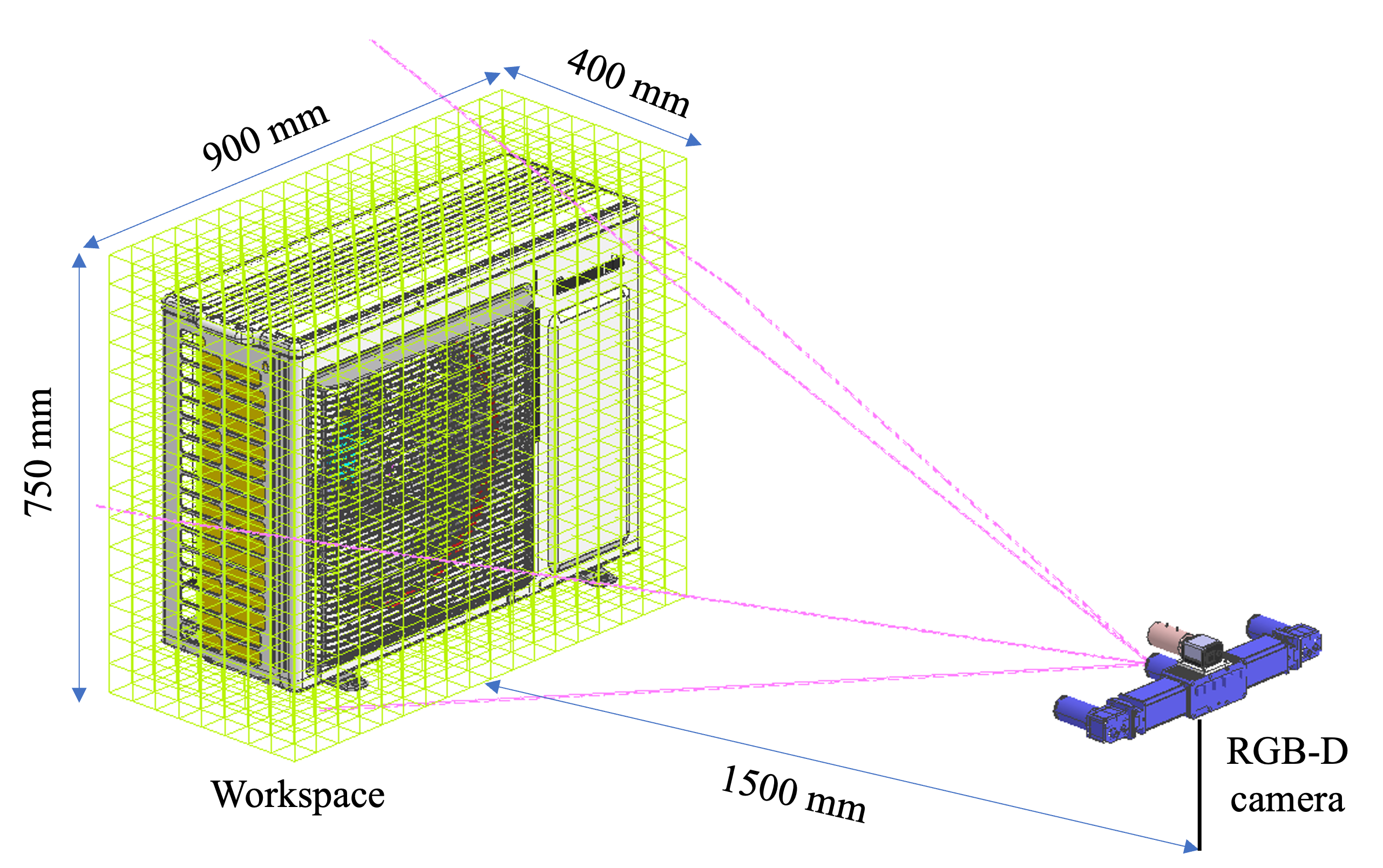}
    \caption{Required manipulation precision of $\pm 0.75$\,mm for screw removal within the workspace ($900 \times 400 \times 750$\,mm, width $\times$ depth $\times$ height).}
    \figlab{proposal}
\end{figure}
\begin{figure}[tb]
    \centering
    \begin{subfigure}[b]{0.48\textwidth}
        \centering
        \includegraphics[width=\textwidth]{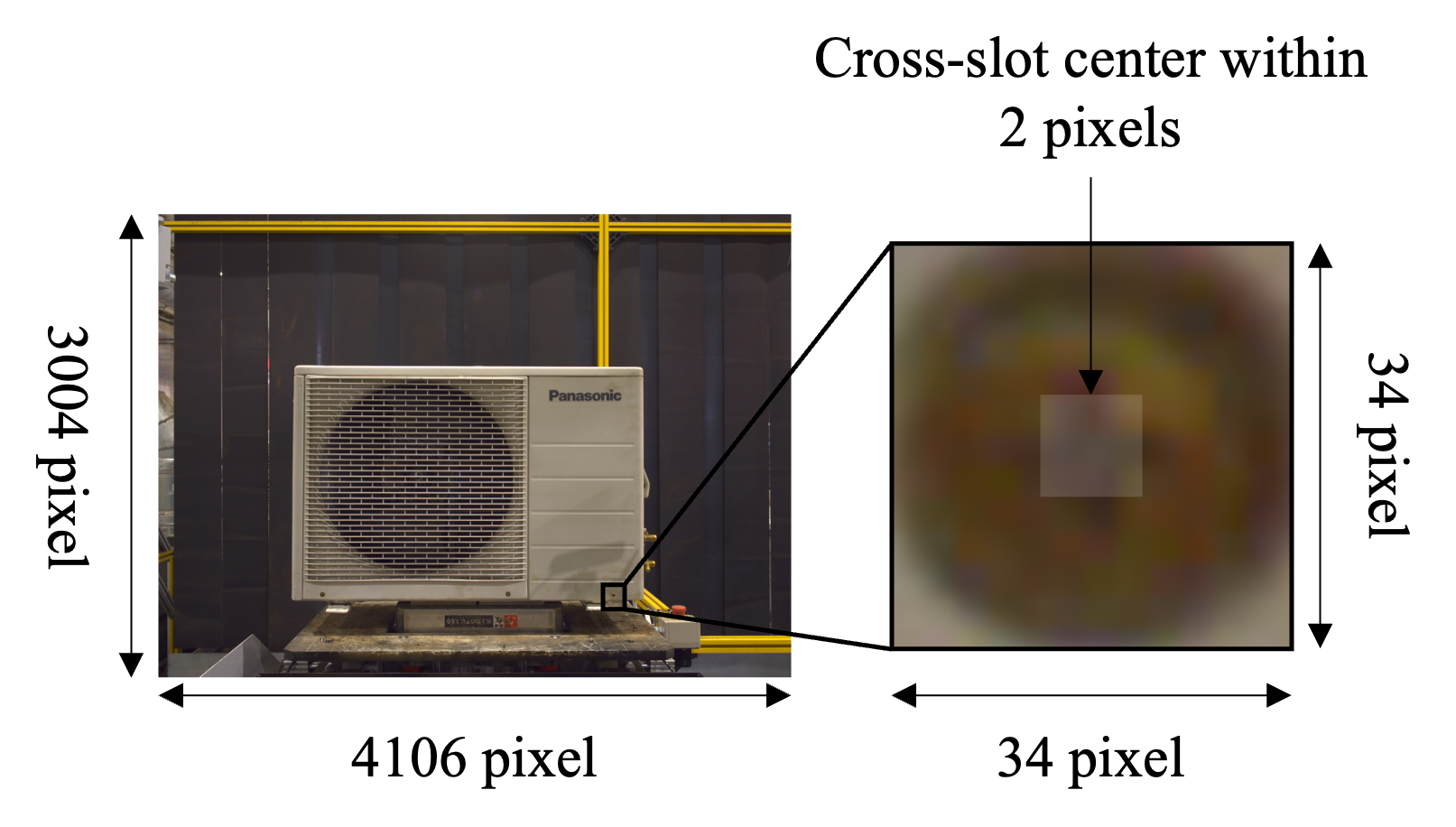}
        \caption{Detected screw area.}
        \label{challenges1}
    \end{subfigure}
    \hfill 
    \begin{subfigure}[b]{0.48\textwidth}
        \centering
        \includegraphics[width=\textwidth]{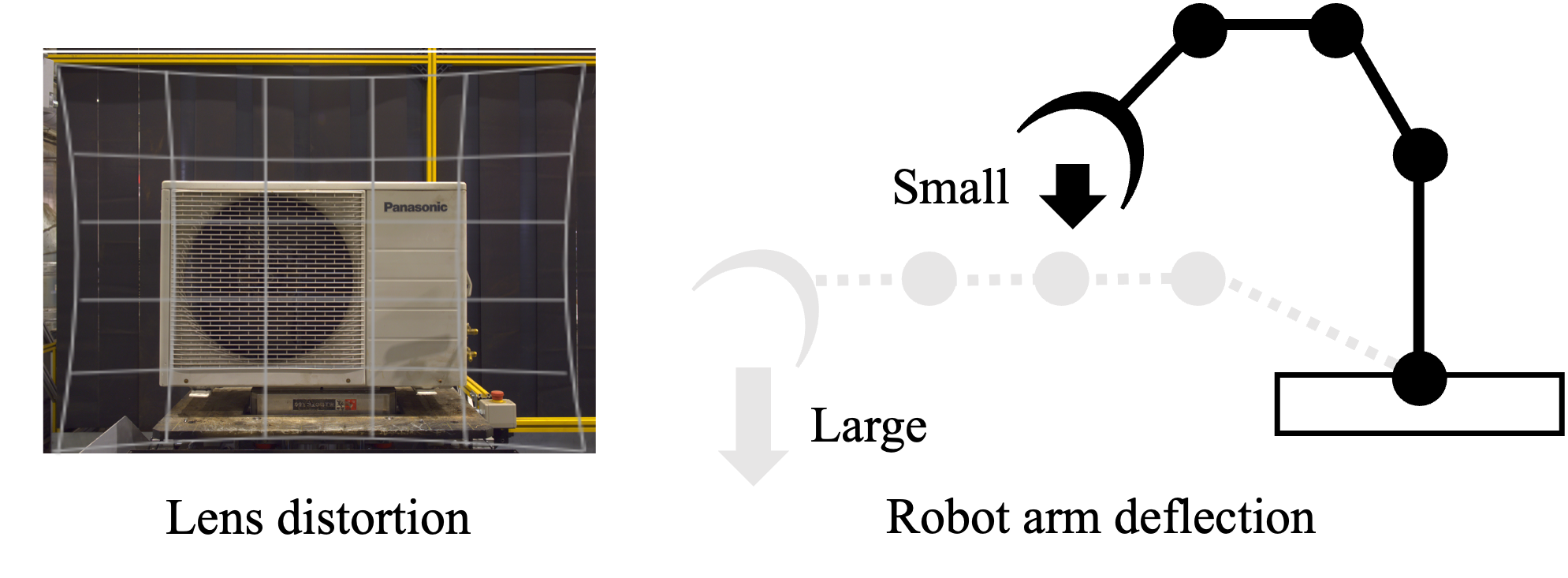} 
        \caption{Sources of manipulation error.}
        \label{challenges2}
    \end{subfigure}
    
    \caption{Associated challenges in the proposed system.}
    \figlab{challenges}
\end{figure}

\subsection{Task-specific two-stage detection for uneven appearances}
\begin{figure}[tb]
    \centering
    \includegraphics[width=0.8\linewidth]{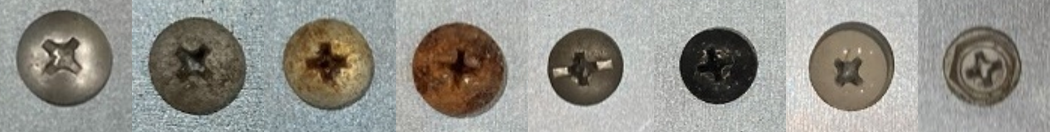}
    \caption{Screws existing in actual recycling factory.}
    \figlab{screws}
\end{figure}
\figref{screws} shows the target screws actually existing in the recycling factory. The challenge of screw detection lies in identifying screws that differ from their original shipment condition, such as those with rust, deformation, or partial damage.
It is essential to detect all approximately 20 screws used in the housing of an air conditioner outdoor unit to accomplish complete the housing disassembly. To achieve the target screw detection rate of 90\%, it is necessary that the overall detection rate satisfies the following relationship: the overall screw detection rate equals the single-screw detection rate raised to the power of 20. From this relationship, the detection rate for each individual screw must exceed 99.5\%.  

In general, screw detection methods such as corner detection and feature point detection can be considered; however, because the target screws exhibit significant variability, stable detection cannot be achieved. Therefore, we detect screw positions using a trained deep learning model and determine the screw centers through an image processing algorithm, achieving a center detection accuracy of $\pm$0.4\,mm.  

When using a single deep-learning-based model, misdetection may occur, resulting in false negatives (failing to detect screws) or false positives (mistakenly identifying non-screw objects as screws). False negatives lead to incomplete disassembly of the air conditioner outdoor unit, while false positives cause the robot to move toward non-screw areas, affecting both cycle time and operational safety.
To address this issue, the proposed screw detection system employs a two-stage detection process and some trained models, to detect screws without omission or excess.

\begin{algorithm}[tb]
\caption{Two-Stage Detection Strategy for variable visual conditions}
\label{alg:screw_detection}
\begin{algorithmic}[1]
\Require Input RGB-D image $\mathbf{I}_{rgbd}$
\Ensure Set of screw center positions $\mathbf{P}_{center}$

\State $\mathbf{I}_{rgb}, \mathbf{D} \leftarrow \textsc{Split}(\mathbf{I}_{rgbd})$
\State $\mathbf{P}_{center} \leftarrow \emptyset$

\Statex \textit{// Stage 1: Coarse Detection (Focus on Recall)}
\State $\mathbf{H}_{coarse} \leftarrow \textsc{HighRecallModel}(\mathbf{I}_{rgb})$
\State $\mathcal{B} \leftarrow \textsc{Threshold}(\mathbf{H}_{coarse})$
\State $\mathcal{I} \leftarrow \textsc{ClipImages}(\mathbf{I}_{rgb}, \mathcal{B})$

\Statex \textit{// Stage 2: Fine Detection (Focus on Precision)}
\For{each image clip $\mathbf{I}_{i} \in \mathcal{I}$}
    \State $\mathbf{I}_{i} \leftarrow \textsc{GammaCorrection}(c_{i})$
    
    \State \textit{// Ensemble Detection on Corrected Image}
    \State $\mathbf{h}_{avg} \leftarrow \frac{1}{3} \sum_{k=1}^{3} \textsc{HighPrecisionModel}_k(I_i)$
    \State $\mathbf{M}_{fine} \leftarrow \textsc{Threshold}(\mathbf{h}_{avg})$
    
    \If{$\textsc{IsScrewDetected}(\mathbf{M}_{fine})$}
        \State \textit{// High-Precision Center Estimation}
        \State $\mathbf{I}_{gray} \leftarrow \textsc{Grayscale}(\mathbf{I}_{i})$
        \State $\mathbf{I}_{eq} \leftarrow \textsc{EqualizeHist}(\mathbf{I}_{gray})$
        \State $\mathbf{p}_{head} \leftarrow \textsc{DetectOuterCircle}(\mathbf{I}_{eq})$ 
        \State $\mathbf{p}_{cross} \leftarrow \textsc{DetectCrossEdge}(\mathbf{I}_{eq})$ 
        \State $\mathbf{p}_{local} \leftarrow \textsc{FuseAndEstimate}(\mathbf{p}_{head}, \mathbf{p}_{cross})$
        
        \State $\mathbf{p}_{global} \leftarrow \textsc{MapToGlobal}(\mathbf{p}_{local}, \mathcal{B})$
        \State $\mathbf{p}_{3d} \leftarrow \textsc{Get3DCoord}(\mathbf{p}_{global}, \mathbf{D})$
        \State $\mathbf{P}_{center} \leftarrow \mathbf{P}_{center} \cup \{ \mathbf{p}_{3d} \}$
    \EndIf
\EndFor

\State \textbf{return} $\mathbf{P}_{center}$
\end{algorithmic}
\end{algorithm}
\begin{figure}[tb]
    \centering
    \includegraphics[width=0.8\linewidth]{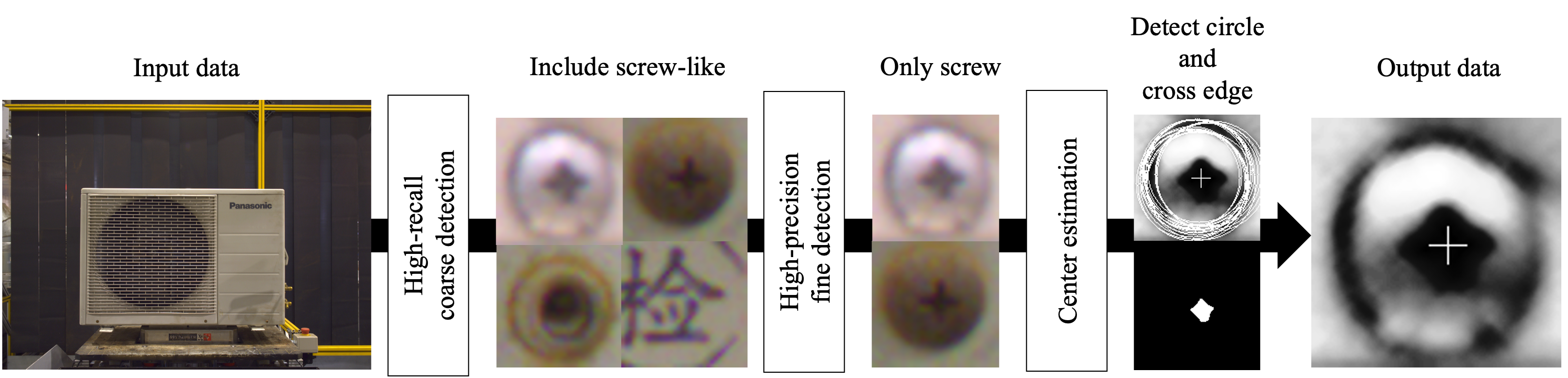}
    \caption{Proposed method for detecting screws with variable visual conditions.}
    \figlab{method}
\end{figure}
Algorithm 1 and \figref{method} outline the proposed method for robust screw detection and high-precision center estimation under uneven lighting conditions. Performance is evaluated based on True Positives (TP), denoting correctly identified screws, and True Negatives (TN), representing the accurate rejection of non-screw features. The system is designed to maximize the TP rate while minimizing False Negatives (FN) and False Positives (FP) through a two-stage detection scheme with an ensemble-based re-verification.

Both stages employ a Fully Convolutional Network (FCN) that outputs probability maps representing the likelihood of screw and non-screw regions. The models are trained in a supervised manner using positional annotations, with a dataset consisting of approximately 39,000 screw images and 43,000 non-screw images, covering diverse screw types, backgrounds, and illumination conditions. Cross-entropy loss is used for training, and the models are optimized with Adam (batch size 128) for 30 epochs. The learning rate is initialized to 0.01 and decayed by a factor of 0.25 every 10 epochs.

In the initialization phase, the input RGB-D image $\mathbf{I}_{rgbd}$ is decomposed by the \textsc{Split} function into the color image $\mathbf{I}_{rgb}$ and the depth map $\mathbf{D}$, while the set of center positions $\mathbf{P}_{center}$ is initialized as empty.
Stage 1 performs coarse detection with a focus on minimizing FN.
First, the \textsc{HighRecallModel} is applied to $\mathbf{I}_{rgb}$ to generate a probability map $\mathbf{H}_{coarse}$.
To suppress noise, this probability map is converted into a binary mask $\mathcal{B}$ by applying the \textsc{Threshold} function and an area filter to retain components exceeding the expected size of a fastener.
Finally, the \textsc{ClipImages} function extracts the set of candidate image patches $\mathcal{I}$ from $\mathbf{I}_{rgb}$ based on the regions defined in $\mathcal{B}$.

Stage 2 performs fine detection with a focus on minimizing FP.
To normalize illumination variance, each candidate image $\mathbf{I}_{i}$ in $\mathcal{I}$ is first pre-processed via the \textsc{GammaCorrection} function.
Subsequently, an ensemble of three models ($\textsc{HighPrecisionModel}_k$) computes an average probability map $\mathbf{h}_{avg}$, which is then binarized into a fine mask $\mathbf{M}_{fine}$ via the \textsc{Threshold} function.
The \textsc{IsScrewDetected} function verifies the candidate based on $\mathbf{M}_{fine}$.

If the candidate is confirmed, the process advances to high-precision center estimation.
First, the patch $\mathbf{I}_{i}$ is converted to grayscale and processed via \textsc{EqualizeHist} to enhance contrast, yielding $\mathbf{I}_{eq}$.
From this enhanced image, two key features are extracted: the screw head center $\mathbf{p}_{head}$ (estimated via circle detection using \textsc{DetectOuterCircle}) and the cross-recess center $\mathbf{p}_{cross}$ (identified via edge and intensity cues using \textsc{DetectCrossEdge}).
The \textsc{FuseAndEstimate} function then integrates these geometric and photometric features to determine the precise local center $\mathbf{p}_{local}$.
Finally, $\mathbf{p}_{local}$ is transformed into the global coordinate $\mathbf{p}_{global}$ within the coordinate system of the original input RGB-D image $I_{rgbd}$ via \textsc{MapToGlobal}. Using this $\mathbf{p}_{global}$, the corresponding depth is retrieved from $\mathbf{D}$ via \textsc{Get3DCoord} to determine the final 3D position $\mathbf{p}_{3d}$, which is then added to $\mathbf{P}_{center}$.
\begin{figure}[tb]
    \centering
    \begin{subfigure}[b]{0.48\textwidth}
        \centering
        \includegraphics[width=\textwidth]{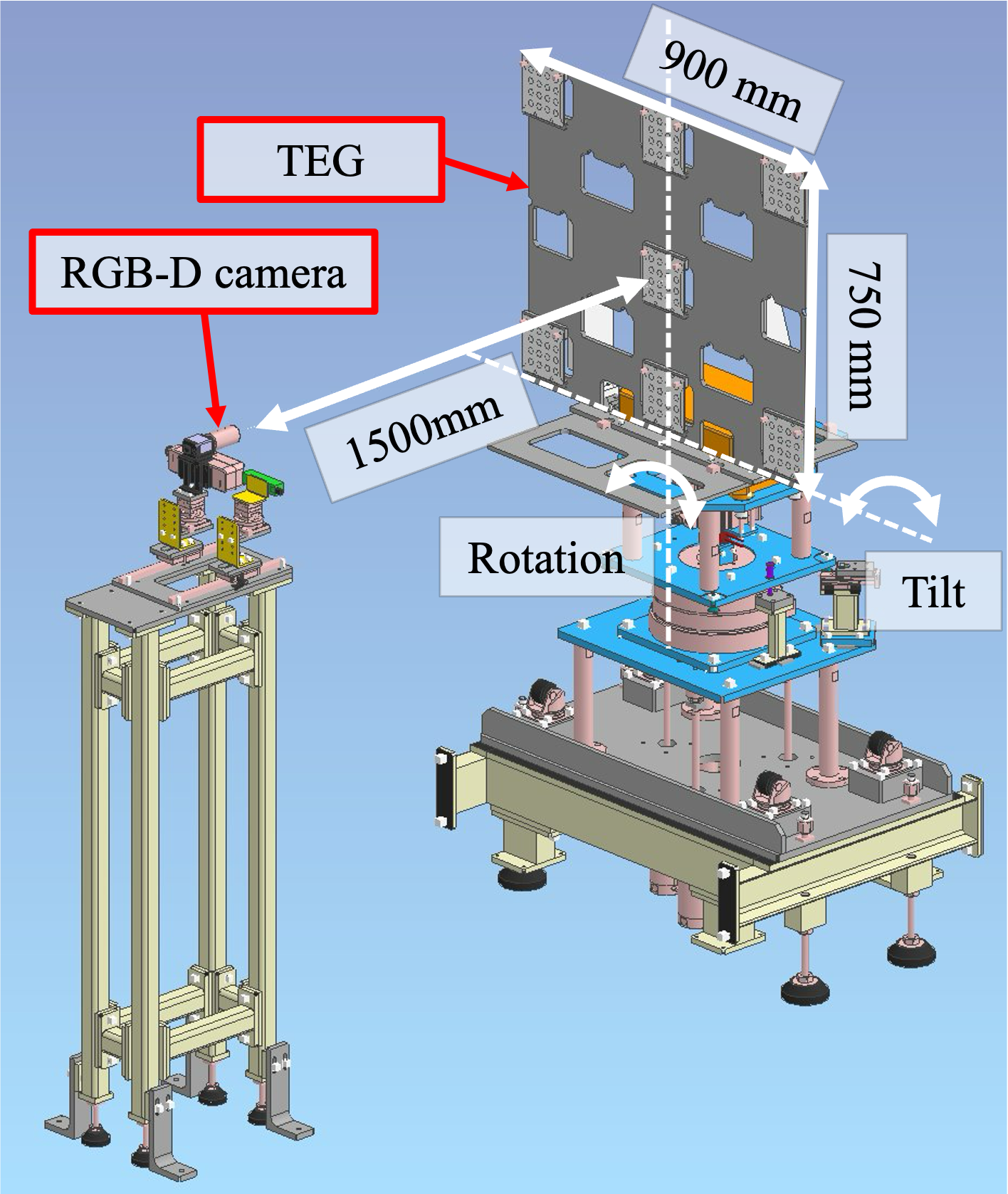}
        \caption{Overview of the TEG.}
        \label{TEG1}
    \end{subfigure}
    \hfill 
    \begin{subfigure}[b]{0.48\textwidth}
        \centering
        \includegraphics[width=\textwidth]{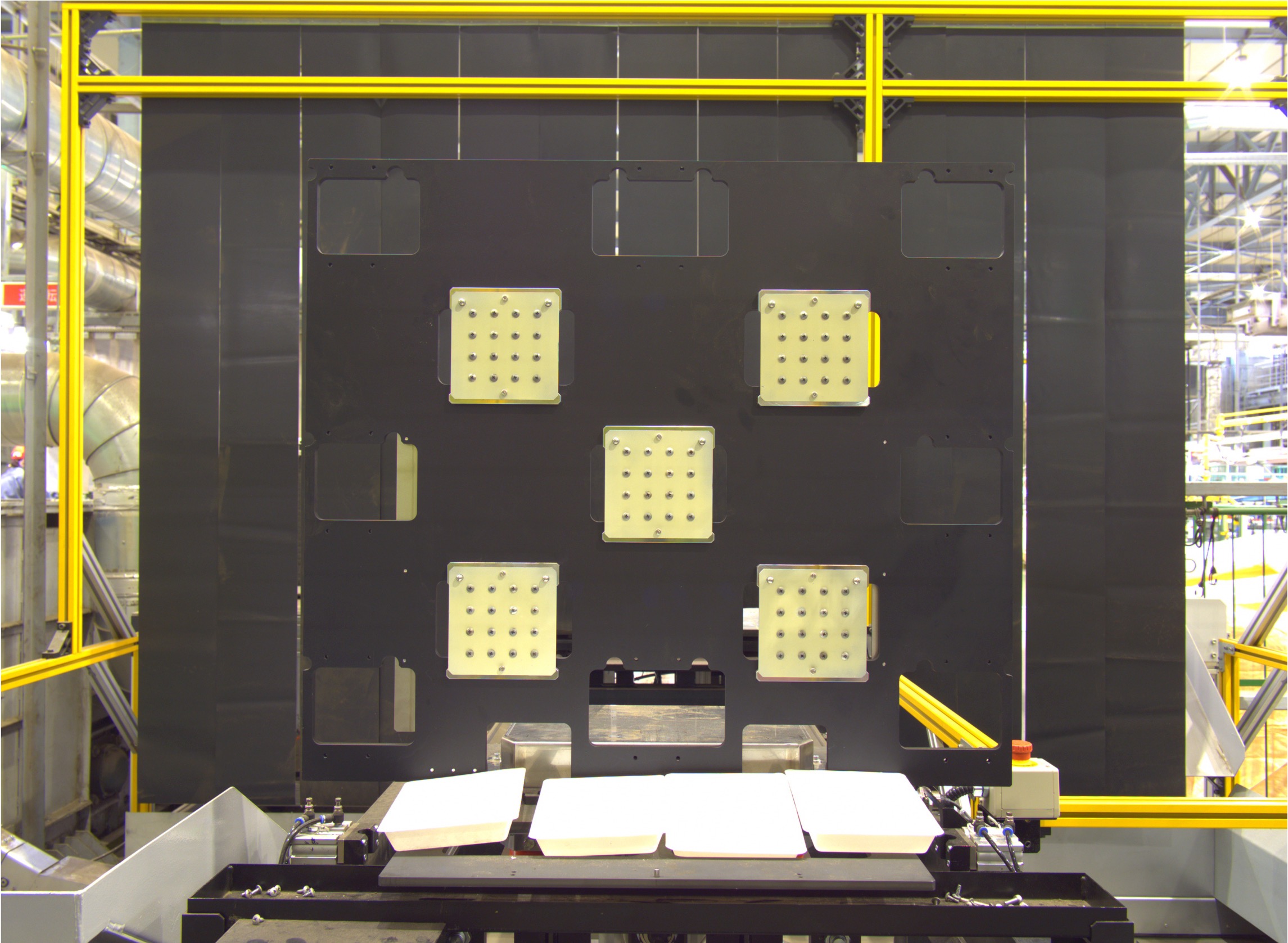} 
        \caption{The TEG captured by the camera.}
        \label{TEG_image}
    \end{subfigure}
    
    \caption{Test Evaluation Gauge (TEG) used for the preliminary evaluation. The gauge replicates the size of an air conditioner outdoor unit and provides controlled pose variations (in-plane rotation $\pm4^\circ$ and tilt $\pm6^\circ$).}
    \figlab{teg}
\end{figure}

\begin{figure}[tb]
    \centering
    \includegraphics[width=0.8\linewidth]{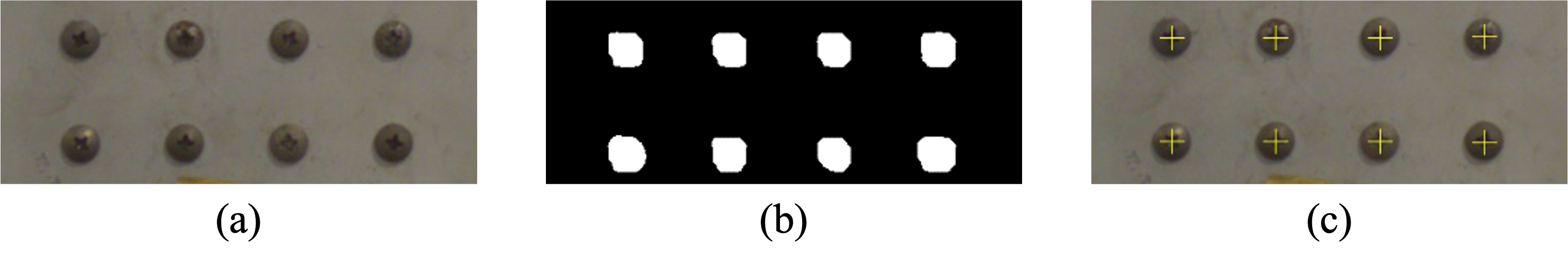}
    \caption{Example of the screw detection process. (a) Input RGB-D image. (b) Probability map. (c) Detection result (yellow markers indicate recognized screw centers).}
    \figlab{result}
\end{figure}

To verify screw detection accuracy, a preliminary experiment was conducted using a Test Evaluation Gauge (TEG). 
The TEG was designed to be equivalent in size to an air conditioner outdoor unit and to include variations within the assumed specification range: $\pm4^\circ$ for in-plane rotation and $\pm6^\circ$ for tilt. 
A total of 3,070 screws were evaluated. The results were classified as follows: 3,064 TP, 6 FN, and 0 FP. 
TN are not applicable as the dataset consisted exclusively of positive samples. Consequently, the method achieved a recall of 99.8\% and a precision of 100.0\%. 
\figref{teg} shows the TEG setup, and \figref{result} presents an example detection result.

\subsection{Precise calibration for industrial-grade robotic screw removal}
\begin{figure}[tb]
    \centering
    \begin{subfigure}[b]{0.48\textwidth}
        \centering
        \includegraphics[width=\textwidth]{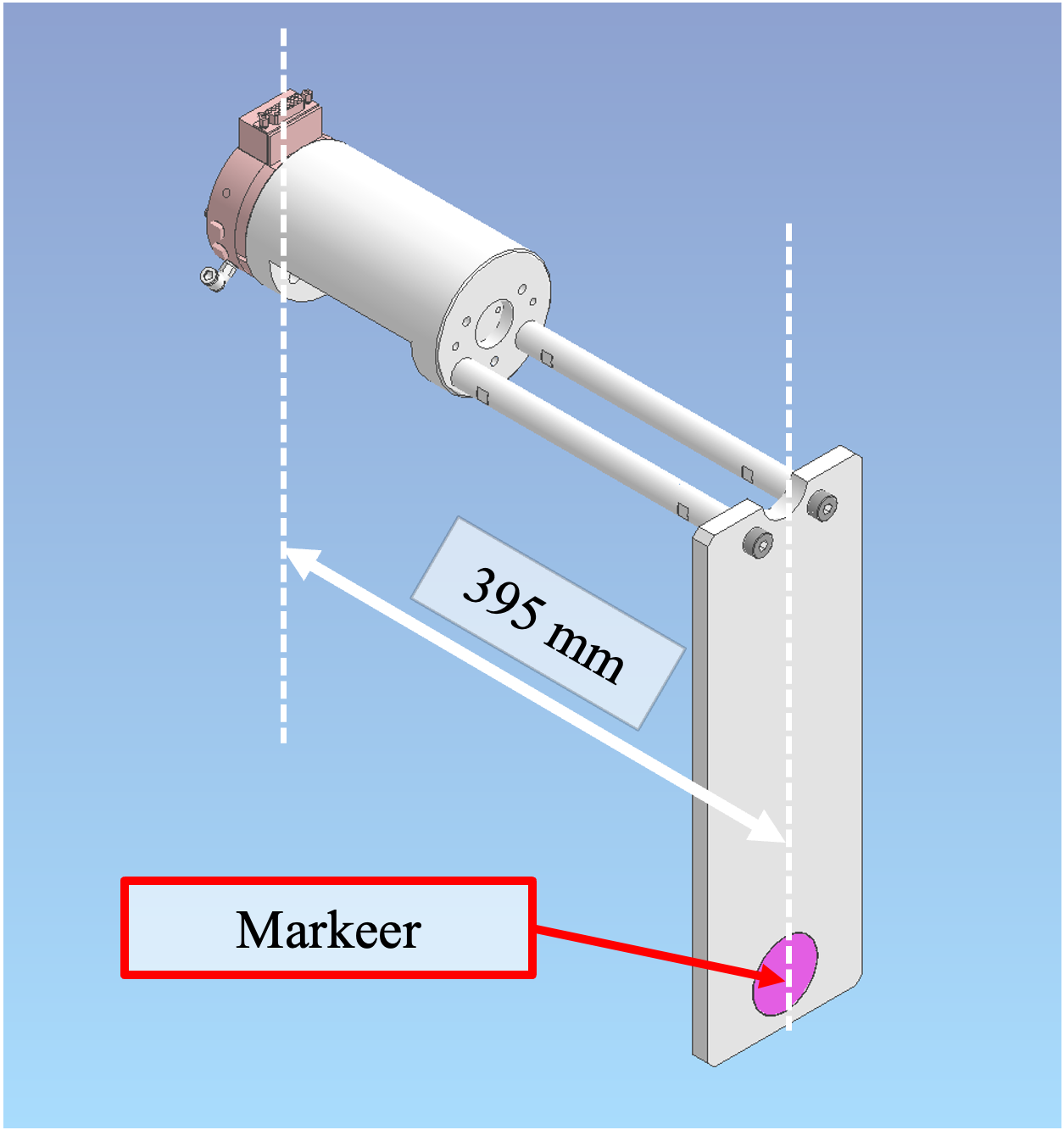}
        \caption{Calibration jig.}
        \label{Calibration jig}
    \end{subfigure}
    \hfill 
    \begin{subfigure}[b]{0.48\textwidth}
        \centering
        \includegraphics[width=\textwidth]{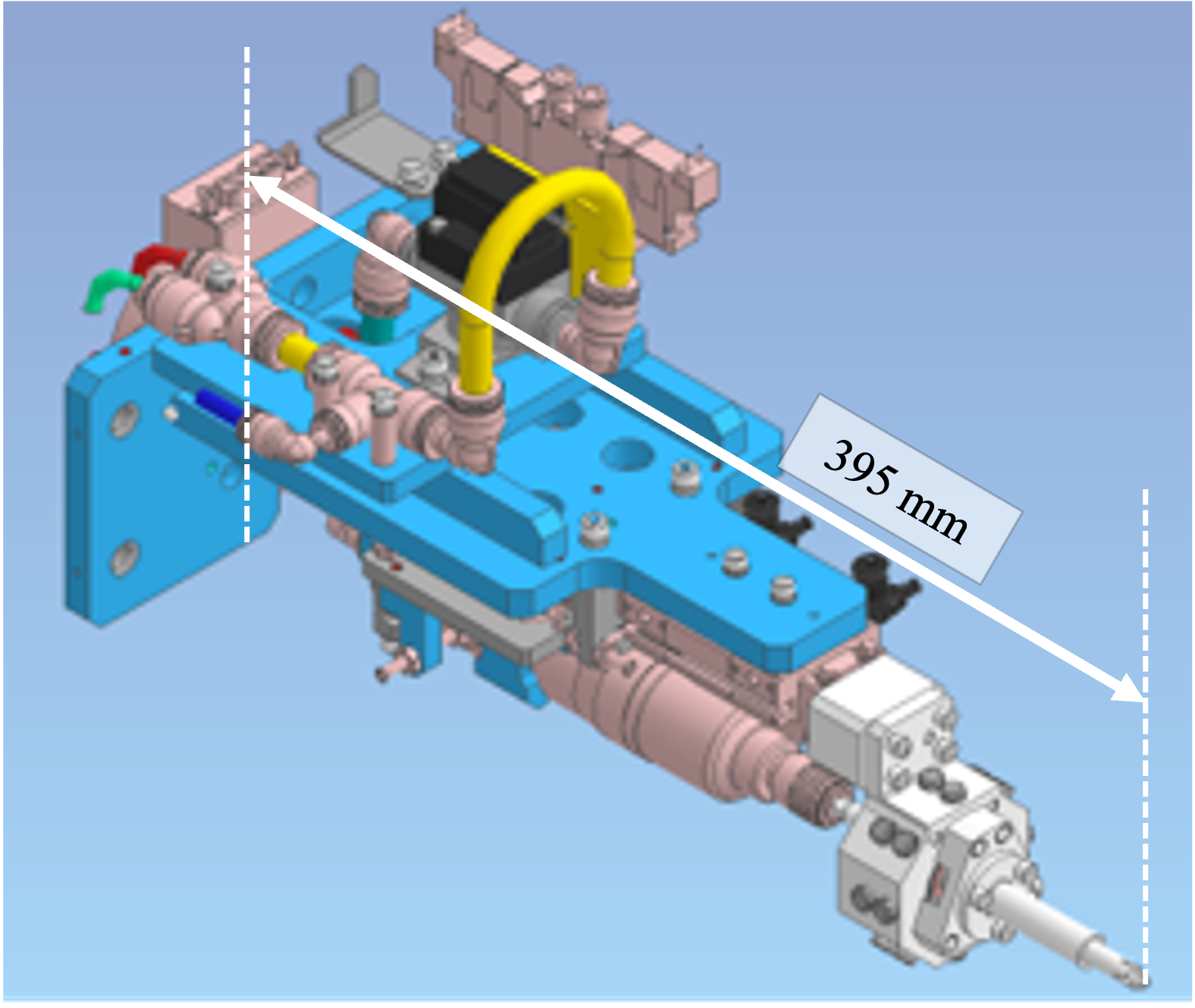} 
        \caption{Driver-hand.}
        \label{Driver-hand}
    \end{subfigure}
    
    \caption{The calibration jig and the driver-hand.}
    \figlab{calibration_jig}
\end{figure}
\begin{figure}[tb]
    \centering
    \includegraphics[width=0.8\linewidth]{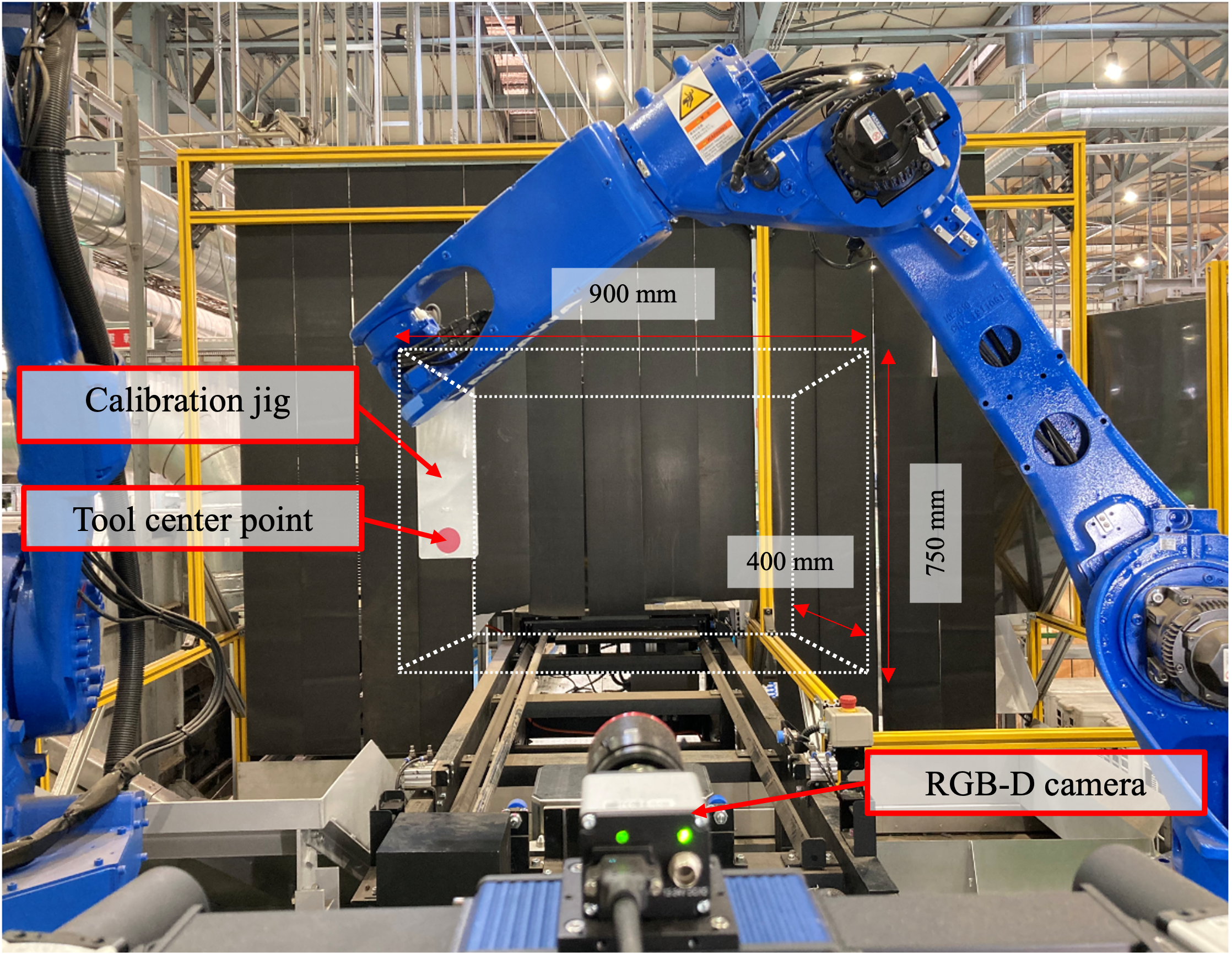}
    \caption{A scene of calibration.}
    \figlab{calibration_scene}
\end{figure}
\begin{figure}[tb]
    \centering
    \includegraphics[width=0.8\linewidth]{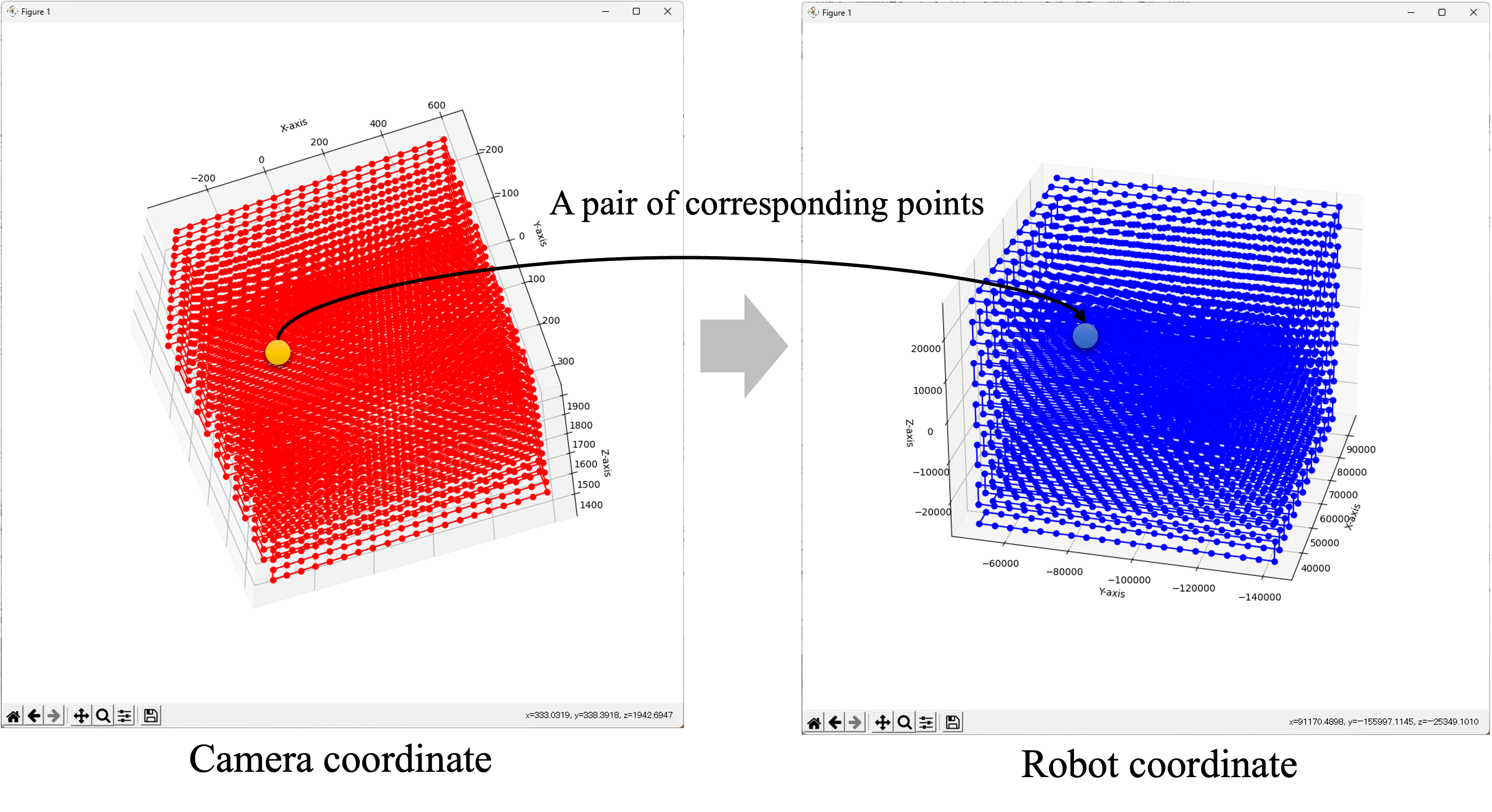}
    \caption{The acquired calibration dataset.}
    \figlab{calibration_dataset}
\end{figure}
To achieve reliable screw removal, the robot must accurately operate at any position detected within the wide workspace observed by the wide-view RGB-D camera, while maintaining a per-screw removal success rate above 99.5\%. This requirement translates into a calibration accuracy such that the discrepancy between RGB-D camera-based position estimates and robot control coordinates remains within $\pm 0.35\,\mathrm{mm}$.

In camera-based picking systems, calibration is commonly performed by estimating a single global rigid transformation (homogeneous transformation) from the camera coordinate system to the robot coordinate system. Using this transformation together with depth information, a 3D point observed by the camera can be mapped to robot coordinates. However, a single global transformation cannot adequately compensate for spatially varying, nonlinear error sources such as residual lens distortion, robot structural deflection, and increased sensitivity near kinematic singularities. Consequently, the position estimation error becomes strongly position-dependent, making it difficult to meet the required manipulation accuracy of $\pm 0.35\,\mathrm{mm}$. Therefore, we adopt a calibration strategy that partitions the workspace into small regions and applies local interpolation.

First, we prepared a calibration jig consisting of a paddle-shaped plate with a single circular marker. The jig mass and the marker offset relative to the robot flange frame were designed to match those of the actual end-effector, the driver-hand, in the $x$ and $y$ directions, while the $z$-direction offset was adjusted to improve marker visibility and detection robustness. The detailed jig dimensions and marker placement are shown in \figref{calibration_jig}.

The driver-hand was developed based on designs from related literature regarding error-compensating end-effectors~\cite{Zhang2023,al2024automated}. It incorporates physical compensation mechanisms to mitigate localization and control errors. Specifically, the integration of elastic components, such as springs, enables the screwdriver bit to passively align with the screw head upon contact, facilitating a self-centering effect. Furthermore, a conical guide is implemented around the bit to physically funnel the screw head into the correct position. To account for seized fasteners, a pneumatic impact driver is adopted. These design features collectively aim to enhance the robustness of the robotic process against sensing and motion inaccuracies, as well as the mechanical resistance of stuck fasteners.

Next, we defined the calibration region as a cuboid in the robot base coordinate system such that it fully covers the target workspace (e.g., an air conditioner outdoor unit). 
 \figref{calibration_scene} shows a scene of the calibration in the real world.
Across a calibration volume of $900\,\mathrm{mm} \times 400\,\mathrm{mm} \times 750\,\mathrm{mm}$  (width $\times$ depth $\times$ height), a 3D lattice of grid points was generated with a $50\,\mathrm{mm}$ spacing, yielding a total of $2{,}736$ points.
As shown in \figref{calibration_scene}, the robot tool center point was sequentially moved to each predefined grid point, and an image was captured at each pose.
From each image, the circular marker was detected using color and area thresholds, and the marker centroid was extracted in image coordinates. The centroid, together with depth information, was back-projected to obtain the corresponding RGB-D camera coordinate, and paired with the commanded robot coordinate at that grid point. Through this procedure, $2{,}736$ correspondence pairs between RGB-D camera and robot coordinates were collected and used as calibration data for interpolation. Data acquisition was fully automated and required approximately 4.5 hours per robot; this procedure is executed only once during system setup and is not required in daily operation. The acquired calibration dataset is shown in \figref{calibration_dataset}.

\begin{algorithm}[tb]
\caption{To convert an arbitrary RGB-D camera coordinate to a robot coordinate}
\label{alg:coordinate_transformation}
\begin{algorithmic}[1]
\Require Target position in camera coordinates $\mathbf{P}_{center}$
\Require Global calibration parameters $\mathbf{T}_{global}$
\Ensure Refined position in robot coordinates $\mathbf{P}_{robot}$

\Statex \textit{// Stage 1: Coarse Mapping (Global Interpolation)}
\State \textit{// Compute an initial estimate using the global linear model}
\State $\mathbf{p}_{rough} \leftarrow \textsc{GlobalLinearMap}(\mathbf{P}_{center}, \mathbf{T}_{global})$ 

\Statex \textit{// Stage 2: Local Neighborhood Identification}
\State \textit{// Extract 27 neighboring points from the calibration context to capture local nonlinearities}
\State $\mathcal{N}_{27} \leftarrow \textsc{GetLocalPoints}(\mathbf{T}_{global}, \mathbf{p}_{rough})$ 

\Statex \textit{// Stage 3: Fine Mapping (Local Interpolation)}
\State \textit{// Apply localized interpolation using the 27 neighbors to compensate for distortion and deflection}
\State $\mathbf{P}_{robot} \leftarrow \textsc{LocalInterpolate}(\mathbf{P}_{center}, \mathcal{N}_{27})$ 

\State \textbf{return} $\mathbf{P}_{robot}$
\end{algorithmic}
\end{algorithm}

To transform an arbitrary camera coordinate $\mathbf{P}_{center}$ into the robot coordinate $\mathbf{P}_{robot}$, we employ a coarse-to-fine approach as outlined in Algorithm 2. 
Initially, a global mapping $\mathbf{T}_{global}$ is used to compute a coarse estimate $\mathbf{p}_{rough}$, which serves to identify the corresponding local neighborhood.
From the calibration data associated with $\mathbf{T}_{global}$, a $3 \times 3 \times 3$ neighborhood of 27 grid points ($\mathcal{N}_{27}$) is selected around this coarse estimate. 
These points are then used to perform a second-order interpolation to obtain the final $\mathbf{P}_{robot}$.
Compared to a single global transformation, this localized refinement strategy effectively mitigates non-linear errors from lens distortion and robot deflection by compensating for inaccuracies within finely partitioned subregions.

Finally, we verified the obtained calibration using millimeter-grid paper and evaluated the positioning error between commanded target positions and actual reached positions. For this evaluation, the bit tip of the driver-hand was modified to a needle shape, and the graph paper was imaged by the RGB-D camera. For an arbitrary target position specified on the image, the corresponding robot coordinate was computed using the proposed calibration, and the robot was commanded to move to that coordinate. The reached position was visualized by plotting a small dot on the paper with the needle tip. The positioning error was computed as the difference between the commanded target location and the plotted location. If the target could not be reached adequately, an additional correction value was introduced within the calibration space based on either the mapping value of the nearest grid point (nearest neighbor) or the interpolated result. Through these verification and correction steps, the required accuracy of $\pm 0.35\,\mathrm{mm}$ was achieved.

From an implementation perspective, explicitly specifying color thresholds, area thresholds, and a handling flow for abnormal detection cases is important to improve marker detection robustness. In addition, the grid spacing introduces a trade-off between interpolation accuracy and data acquisition cost; thus, the spacing should be determined experimentally.

\section{Real-world experimental evaluations}
\subsection{Overview}
Our experiments verified whether the developed system could achieve the target disassembly completion rate of 75\% or higher and a cycle time of 216 seconds or less for the disassembly of air conditioner outdoor unit housings.
Achieving these performance targets necessitates high-recall detection, requiring a recognition accuracy of at least 99.5\% and a robot motion accuracy of $\pm 0.75\,\mathrm{mm}$, to ensure consistent operation and high reliability.
The verification was conducted using 120 air conditioner outdoor units actually collected at a recycling plant. 
All units had undergone preliminary processing, including refrigerant recovery and the disconnection of power wiring.

For evaluating the disassembly completion rate, each unit was categorized into one of the following three classifications: (I) successfully disassembled (OK), (II) not completely disassembled and discharged (NG), or (III) excluded items that could not be disassembled by the system. The excluded items included severely seized screws, screws whose cross slots were covered by spider webs, screws with mud-filled slots, and screws on deformed covers where the mounting surface was inclined beyond the specified range. For evaluating the cycle time, the measurement was triggered by the loading switch and recorded internally by the computer until the completion of discharge.

\subsection{Target workpieces and disassembly procedure}
\figref{target} illustrates the target outdoor units.
While these units encompass various model numbers, they share a consistent structural configuration consisting of front, left, right, and top panels.
These units are assumed to fit within a spatial envelope of $900 \times 400 \times 750$ mm.
The disassembly process follows a predefined sequence: (1) front cover screw removal, (2) left cover screw removal, (3) right cover screw removal, (4) top cover removal, (5) front inner cover screw removal, (6) front cover removal, and (7) right cover removal.
Note that this sequence represents a high-level workflow. 

The system dynamically detects the precise positions of screws and covers, thereby accommodating a wide variety of air conditioner outdoor unit models. Furthermore, a specific step for removing the left cover is omitted, as this cover detaches simultaneously with the removal of the front cover.
\begin{figure}[tb]
    \centering
    \includegraphics[width=0.8\linewidth]{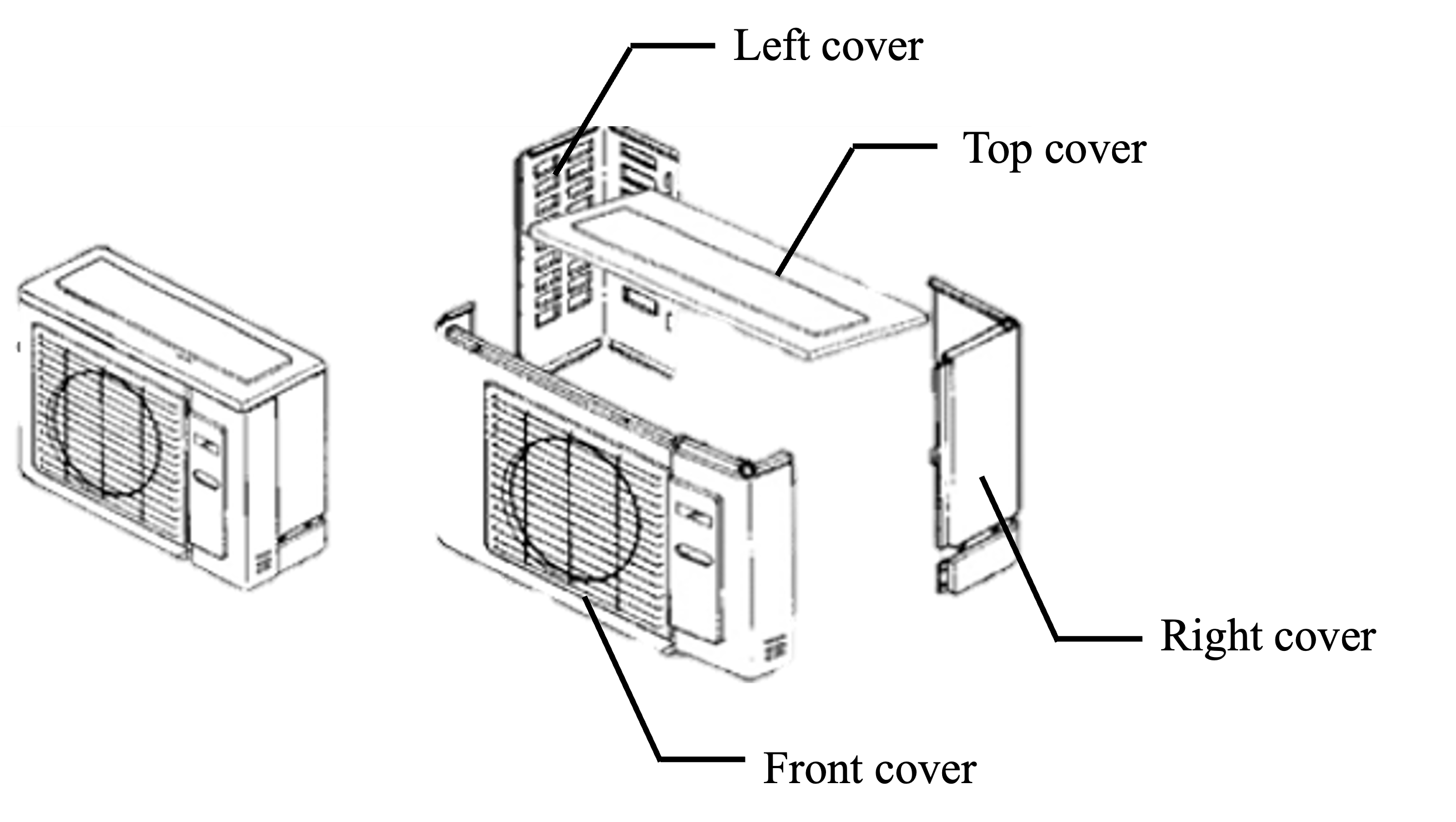}
    \caption{Target workpieces.}
    \figlab{target}
\end{figure}

\subsection{Robotic disassembly results}
\figref{cu_dis} shows the disassembly process. Among the 120 tested units, 72 units were classified as OK, 20 as NG, and 28 as excluded. 
Excluding the 28 units categorized as excluded, the system successfully completed the disassembly of 72 out of 92 units, resulting in a disassembly completion rate of 78.3\%. 
The average disassembly cycle time for OK units was 197 seconds, and for NG units, 191 seconds. 
A detailed analysis of the disassembly cycle time showed that approximately 45 seconds were required for material handling, 80 seconds for screw removal, 45 seconds for cover removal, and 30 seconds for inspection processes.

\figref{exception} illustrates representative examples of the 28 excluded units, which account for approximately 24\% of the 120 units.
As shown in (a) and (b), some screws were heavily occluded by accumulated dirt or spider webs, making visual detection impossible.
Others, depicted in (c) and (d), exhibited severe structural damage, such as deformation or cracking of the housing, which prevented stable manipulation.

Based on these results, the developed system achieved both targets: a disassembly completion rate of 78.3\% and an average cycle time of 193 seconds, satisfying the goals of 75\% or higher and 216 seconds or less, respectively. 
\begin{figure}[tb]
    \centering
    \includegraphics[width=0.8\linewidth]{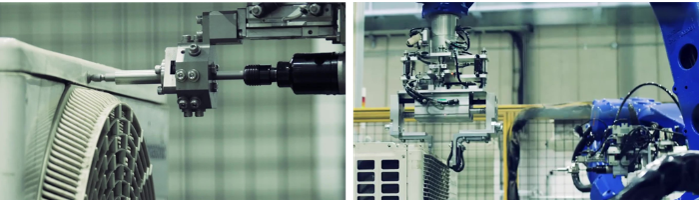}
    \caption{Robotic disassembly for actual air conditioner outdoor unit.}
    \figlab{cu_dis}
\end{figure}
\begin{figure}[tb]
    \centering
    \includegraphics[width=0.8\linewidth]{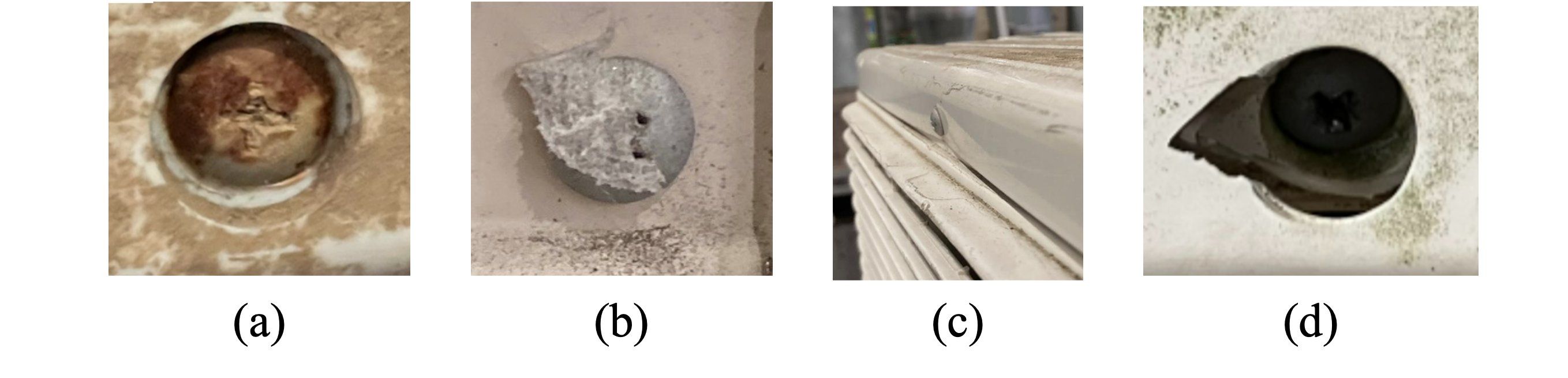}
    \caption{Exceptional detection failure cases: (a, b) dirt/spider webs; (c, d) severe deformation/cracking.}
    \figlab{exception}
\end{figure}
\section{Discussions}
The experimental results demonstrate that the system successfully met all performance targets, suggesting the overall effectiveness of the proposed method. Specifically, the developed system achieved a disassembly completion rate of 78.3\% and an average cycle time of 193 seconds, satisfying the predefined goals of $> 75\%$ and $\le 216$  seconds, respectively.
However, to realize even faster processing, higher success rates, and broader versatility, we believe it is necessary to advance beyond our system. 
We propose four key directions for future development: Design for Circular Economy (DfCE), automated process planning, a Human-in-the-Loop system, and detection-less controlled destructive disassembly.

\subsection{Design for Circular Economy (DfCE)}
To accelerate resource circulation, optimization must occur not only at the recycling plant but also during the product design phase. We advocate for "Robot-Friendly Design" within the framework of DfCE. This involves redesigning screw layouts, reducing screw counts, and simplifying cover structures to facilitate robotic access~\cite{Xu2025EcoDesign}. 

Specifically, by adopting structures that are easier for vision systems to detect and for end-effectors to access regardless of deformation or heavy soiling, we can significantly reduce the number of "NG" and "Excluded" units.
This upstream approach is essential for realizing robust, high-speed disassembly.

\subsection{Generalization and process optimization}
Although this study focused on air conditioner outdoor units, actual recycling plants handle a diverse range of appliances, including televisions, washing machines, refrigerators, and microwave ovens. 
The proposed technology has the potential to be expanded to these domains; for instance, \figref{mo_dis} suggests its applicability to screw and cover removal in microwave ovens. 

A current bottleneck is that rough disassembly flows are generated manually. Creating manual workflows for such a vast variety of appliances is inefficient and unscalable. Furthermore, the current process relies on a single robot with potentially redundant motion sequences. To address this, future systems should utilize design data (such as CAD) to automatically generate optimal disassembly flows and employ multi-robot task scheduling. 
These capabilities are currently being developed in a separate study~\cite{Kiyokawa2026schedule} and are planned to be integrated into this system in the near future.
This would eliminate redundancy and enable parallel processing, leading to significantly faster and more robust operations.

\subsection{Human-in-the-Loop for continuous learning}
In this study, we propose a Human-in-the-Loop approach to handle the inherent uncertainties of recycling. Since no two waste products are in the identical condition, deep-learning-based detection failures are inevitable. A system is required where human operators can intervene to correct these failures during the disassembly task. Crucially, this intervention should not just be a fix, but a data collection process. 

Rather than relying on offline annotation, the proposed system is designed to accumulate 'live' training data during actual plant operation.
This continuous feedback loop will allow the deep learning model to adapt to rare cases and improve its accuracy over time, establishing a sustainable cycle of automation and learning.

\subsection{Detection-less controlled destructive disassembly}
To further enhance throughput and accommodate units that are currently excluded, we believe that detection-less controlled destructive disassembly is essential. The necessity of this approach is underscored by the fact that approximately 23.3\% of the tested units (28 out of 120) were unsuitable for the current vision-based method. As detailed in the previous section, these units present fundamental challenges for visual detection and stable mechanical manipulation.

To overcome these limitations, it is necessary to explore alternative disassembly strategies, such as the drilling and milling techniques introduced by Al Assadi et al.~\cite{al2024automated} . By utilizing mechanical force to bypass the requirement for precise screw localization, the system can potentially process even highly degraded or damaged units that are otherwise unmanageable. However, it is important to note a critical trade-off: while these destructive methods are highly effective for material recycling, they render the non-destructive recovery of components impossible. Consequently, their application must be strategically managed and selected based on whether the priority is component reuse or material recovery within the circular economy.
\begin{figure}[tb]
    \centering
    \includegraphics[width=0.8\linewidth]{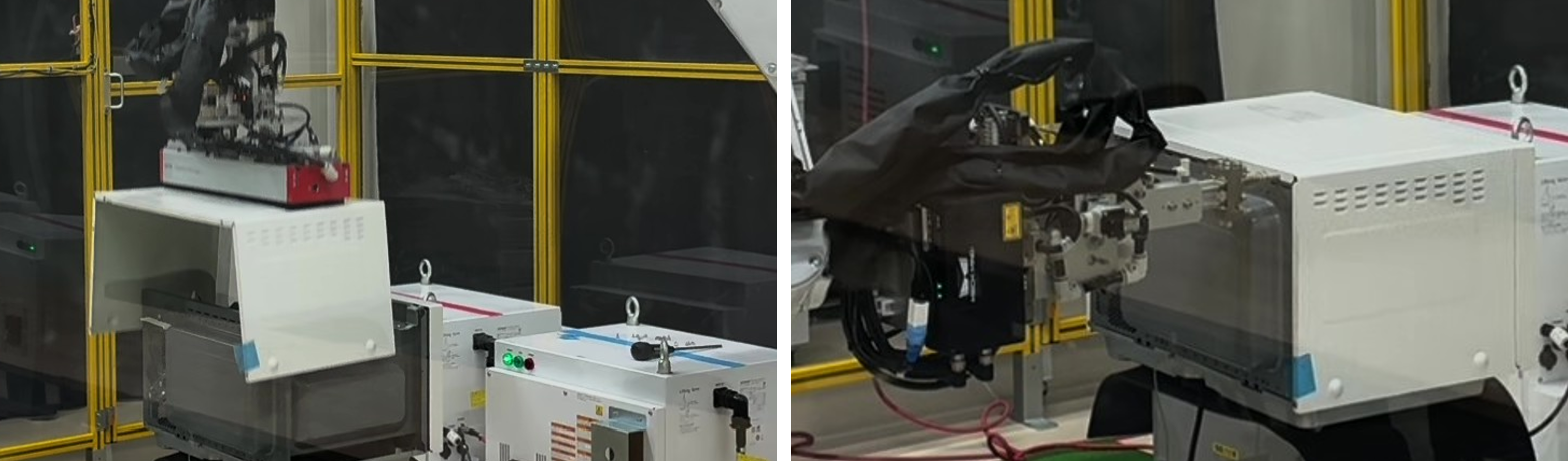}
    \caption{Robotic disassembly for actual microwave oven.}
    \figlab{mo_dis}
\end{figure}

\section{Conclusion}



In this study, we aimed to develop an automated system for the reversible disassembly of air conditioner outdoor units, specifically targeting the removal of fasteners and exterior housing. To achieve industrial viability, we established performance targets of a disassembly completion rate exceeding 75\% and a cycle time of less than 216 seconds.

First, to address the challenge of detecting fasteners across diverse and degraded models, we proposed a task-specific two-stage detection method.
By combining coarse estimation for high recall with ensemble-based fine verification, our approach achieved robust identification of screws with variable visual conditions, such as rust or deformation. In an evaluation of 3,070 screws, the method demonstrated a recall of 99.8\% and a precision of 100.0\% (3,064 TP, 0 FP), realizing the high-recall detection required for downstream processes.

Second, to ensure robust manipulation, we implemented a lattice-based local calibration strategy. This technique effectively compensated for spatially varying nonlinear errors, specifically residual lens distortion and robot structural deflection.
Consequently, the system successfully achieved the critical manipulation accuracy target of $\pm$0.75 mm using only a single RGB-D camera, without relying on pre-programmed coordinates.

Finally, the system's practical feasibility was demonstrated through extensive real-world experiments involving 120 units sourced from an actual recycling plant.
The system achieved a disassembly success rate of 78.3\% and an average cycle time of 193 seconds. Both metrics surpassed the initial development targets, confirming the effectiveness of the proposed technologies for industrial applications.

\section*{Disclosure statement}
No potential conflict of interest was reported by the authors.

\section*{Funding}
This article is based on results obtained from the project JPNP23002, commissioned by the New Energy and Industrial Technology Development Organization (NEDO).

\section*{Notes on contributors}
\noindent\textit{\textbf{Tomoki Ishikura}} obtained his master's degree from Nara Institute of Science and Technology in 2021. Since joining the company, he has been engaged in the development of an automated disassembly system for used home appliances at a recycling facility. He is currently pursuing the Ph.D. degree in the Graduate School of Engineering Science, the University of Osaka, Toyonaka, Japan. Recently, he has focused on refurbishment as a circular economy business and is working on developing an autonomous disassembly system using AI and robotics technology to revolutionize the resource circulation system centered on crushing.
\vspace{2mm}

\noindent\textit{\textbf{Genichiro Matsuda}} is a manager of resource circulation technology development in the R\&D department of Panasonic Holdings. He has been in charge of the development of home appliance recycling technology since 2015. Toward the realization of a sustainable society, he has promoted the development of recycling technology for cellulose composite plastics and the development of resource efficiency indicators for circular economy products and services. In recent years, he has focused on refurbishment as a circular economy business, and is developing an autonomous disassembly system that utilizes AI and robotics technology to revolutionize the resource circulation system centered on crushing.
\vspace{2mm}

\noindent\textit{\textbf{Takuya Kiyokawa}} received the B.E. degree from the National Institute of Technology, Kumamoto College, Japan, and the M.E. degree and the Ph.D. degree in engineering from the Nara Institute of Science and Technology, Japan, in 2018 and 2021, respectively. From 2021 to 2022, he was with the University of Osaka, Japan, as a Specially-Appointed Assistant Professor, and with the Nara Institute of Science and Technology, as a Specially-Appointed Assistant Professor. From 2023 to 2024, he was a Visiting Researcher with the Institute of Robotics and Mechatronics, German Aerospace Center (DLR), Oberpfaffenhofen, Weßling, Germany and since 2023, he has been with the University of Osaka, as an Assistant Professor. His current research interests include robot manipulation and agile reconfigurable robotic systems.
\vspace{2mm}

\noindent\textit{\textbf{Kensuke Harada}} received the B.Sc., M.Sc., and Ph.D. degrees in mechanical engineering from Kyoto University, Kyoto, Japan, in 1992, 1994, and 1997, respectively. From 1997 to 2002, he was a Research Associate with Hiroshima University, Hiroshima, Japan. Since 2002, he has been with the National Institute of Advanced Industrial Science and Technology (AIST). From 2005 to 2006, he was a Visiting Scholar with the Department of Computer Science, Stanford University, Stanford, CA, USA, and the Leader of the Manipulation Research Group, AIST, from 2013 to 2015. He is currently a Professor with the Graduate School of Engineering Science, the University of Osaka, Toyonaka, Japan. His research interests include mechanics and control of robot manipulators and robot hands, biped locomotion, and motion planning of robotic systems.

\bibliographystyle{tfnlm}
\bibliography{interactnlmsample}

@techreport{Balde2024GEM,
  title        = {Global E-waste Monitor 2024},
  author       = {Bald{\'e}, Cornelis P. and Kuehr, Ruediger and Yamamoto, Tales and McDonald, Rosie and D'Angelo, Elena and Althaf, Shahana and Bel, Garam and Deubzer, Otmar and Fernandez-Cubillo, Elena and Forti, Vanessa and Gray, Vanessa and Herat, Sunil and Honda, Shunichi and Iattoni, Giulia and Khetriwal, Deepali S. and Luda di Cortemiglia, Vittoria and Lobuntsova, Yuliya and Nnorom, Innocent and Pralat, No{\'e}mie and Wagner, Michelle},
  year         = {2024},
  institution  = {International Telecommunication Union (ITU) and United Nations Institute for Training and Research (UNITAR)},
  address      = {Geneva/Bonn},
}

@techreport{aeha2022report,
  title       = {Annual Report on Home Appliance Recycling 2022},
  author      = {{Association for Electric Home Appliances}},
  institution = {Association for Electric Home Appliances (AEHA)},
  year        = {2023},
  address     = {Tokyo, Japan},
}

@Article{electronics11040533,
AUTHOR = {Cheng, Chen-Yang and Chen, Yin-Yann and Pourhejazy, Pourya and Lee, Chih-Yu},
TITLE = {Disassembly Line Balancing of Electronic Waste Considering the Degree of Task Correlation},
JOURNAL = {Electronics},
VOLUME = {11},
YEAR = {2022},
NUMBER = {4},
ARTICLE-NUMBER = {533},
ISSN = {2079-9292},
DOI = {10.3390/electronics11040533}
}

@article{HONG2015357,
title = {Life cycle assessment of electronic waste treatment},
journal = {Waste Management},
volume = {38},
pages = {357-365},
year = {2015},
issn = {0956-053X},
doi = {https://doi.org/10.1016/j.wasman.2014.12.022},
author = {Jinglan Hong and Wenxiao Shi and Yutao Wang and Wei Chen and Xiangzhi Li}
}

@article{MENIKPURA2014183,
title = {Assessing the climate co-benefits from Waste Electrical and Electronic Equipment (WEEE) recycling in Japan},
journal = {Journal of Cleaner Production},
volume = {74},
pages = {183-190},
year = {2014},
issn = {0959-6526},
doi = {https://doi.org/10.1016/j.jclepro.2014.03.040},
url = {https://www.sciencedirect.com/science/article/pii/S0959652614002613},
author = {S.N.M. Menikpura and Atsushi Santo and Yasuhiko Hotta}
}

@article{RAVI2012145,
title = {Evaluating overall quality of recycling of e-waste from end-of-life computers},
journal = {Journal of Cleaner Production},
volume = {20},
number = {1},
pages = {145-151},
year = {2012},
issn = {0959-6526},
doi = {https://doi.org/10.1016/j.jclepro.2011.08.003},
author = {V. Ravi}
}

@article{ASIF2024483,
title = {Robotic disassembly for end-of-life products focusing on task and motion planning: A comprehensive survey},
journal = {Journal of Manufacturing Systems},
volume = {77},
pages = {483-524},
year = {2024},
issn = {0278-6125},
doi = {https://doi.org/10.1016/j.jmsy.2024.09.010},
author = {Mohammed Eesa Asif and Alireza Rastegarpanah and Rustam Stolkin}
}

@Article{app15020618,
AUTHOR = {Clark, Austin and Jouaneh, Musa K.},
TITLE = {A System for Robotic Extraction of Fasteners},
JOURNAL = {Applied Sciences},
VOLUME = {15},
YEAR = {2025},
NUMBER = {2},
ARTICLE-NUMBER = {618},
ISSN = {2076-3417},
DOI = {10.3390/app15020618}
}

@Article{app14146301,
AUTHOR = {DiFilippo, Nicholas M. and Jouaneh, Musa K. and Jedson, Alexander D.},
TITLE = {Optimizing Automated Detection of Cross-Recessed Screws in Laptops Using a Neural Network},
JOURNAL = {Applied Sciences},
VOLUME = {14},
YEAR = {2024},
NUMBER = {14},
ARTICLE-NUMBER = {6301},
ISSN = {2076-3417},
DOI = {10.3390/app14146301}
}

@inproceedings{Zhang2023,
author = {Zhang, Yisheng and Zhang, Hengwei and Wang, Zhigang and Zhang, Shengmin and Li, Huaicheng and Chen, Ming},
booktitle = {Proc. IEEE/ASME Int. Conf. Adv. Intell. Mech.}, 
title = {Development of an Autonomous, Explainable, Robust Robotic System for Electric Vehicle Battery Disassembly}, 
year = {2023},
volume = {},
number = {},
pages = {409--414}}

@techreport{Balde2022,
  title={Global Transboundary E-waste Flows Monitor 2022},
  author={Bald{\'e}, K. and D'Angelo, E. and Luda, V. and Deubzer, O. and K{\"u}hr, R.},
  url={https://books.google.co.jp/books?id=AavOzwEACAAJ},
  year={2022},
  publisher={United Nations Institute for Training and Research (UNITAR)}
}

@article{UEDA2024,
title = {Automatic high-speed smartphone disassembly system},
journal = {Journal of Cleaner Production},
volume = {434},
pages = {139928},
year = {2024},
issn = {0959-6526},
doi = {https://doi.org/10.1016/j.jclepro.2023.139928},
url = {https://www.sciencedirect.com/science/article/pii/S0959652623040866},
author = {Takao Ueda and Hideaki Fukusawa and Yukimi Nakagawa and Kazuya Nagano and Naoki Sunahara and Hiroshi Yamada and Junichi Fujisawa and Makoto Yamada and Shigeki Koyanaka and Tatsuya Oki}
}

@article{Deng2024,
author = {Wupeng Deng and Quan Liu and Duc Truong Pham and Jiwei Hu and Kin-Man Lam and Yongjing Wang and Zude Zhou},
title = {Predictive exposure control for vision-based robotic disassembly using deep learning and predictive learning},
journal = {Robotics and Computer-Integrated Manufacturing},
volume = {85},
pages = {102619},
year = {2024}}

@article{Diaz2025,
author = {Iñaki Díaz and Diego Borro and Olatz Iparraguirre and Martxel Eizaguirre and Frank A. Ricardo and Nicolás Muñoz and Jorge Juan Gil},
title = {Robotic system for automated disassembly of electronic waste: Unscrewing},
journal = {Robotics and Computer-Integrated Manufacturing},
volume = {95},
pages = {103032},
year = {2025}}

@article{Difilippo2018,
author = {DiFilippo, Nicholas M. and Jouaneh, Musa K.},
journal = {IEEE Transactions on Automation Science and Engineering}, 
title = {A System Combining Force and Vision Sensing for Automated Screw Removal on Laptops}, 
year = {2018},
volume = {15},
number = {2},
pages = {887--895}}

@article{Li2020,
author = {Li, Ruiya and Pham, Duc Truong and Huang, Jun and Tan, Yuegang and Qu, Mo and Wang, Yongjing and Kerin, Mairi and Jiang, Kaiwen and Su, Shizhong and Ji, Chunqian and Liu, Quan and Zhou, Zude},
journal = {IEEE Transactions on Automation Science and Engineering}, 
title = {Unfastening of Hexagonal Headed Screws by a Collaborative Robot}, 
year = {2020},
volume = {17},
number = {3},
pages = {1455--1468}}

@article{Zhang2022,
author = {Zhang, Xinyao and Eltouny, Kareem and Liang, Xiao and Behdad, Sara},
title = {Automatic Screw Detection and Tool Recommendation System for Robotic Disassembly},
journal = {Journal of Manufacturing Science and Engineering},
volume = {145},
number = {3},
pages = {031008},
year = {2022}}

@article{MATSUTO2004425,
title = {Material and heavy metal balance in a recycling facility for home electrical appliances},
journal = {Waste Management},
volume = {24},
number = {5},
pages = {425-436},
year = {2004},
issn = {0956-053X},
doi = {https://doi.org/10.1016/j.wasman.2003.12.002},
url = {https://www.sciencedirect.com/science/article/pii/S0956053X03002320},
author = {T. Matsuto and C.H. Jung and N. Tanaka}
}

@book{Boothroyd2010,
  title={Product Design for Manufacture and Assembly},
  author={Boothroyd, Geoffrey and Dewhurst, Peter and Knight, Winston},
  year={2010},
  publisher={CRC press}
}

@book{Vongbunyong2015,
  author = {Vongbunyong, Supachai and Kara, Sami and Pagnucco, Maurice},
  title = {Automated Systems with Cognitive Abilities},
  publisher = {Springer},
  year = {2015},
  doi = {10.1007/978-3-319-12502-2}
}

@article{Vanegas2018,
  author = {Vanegas, Paul and Peeters, Jef R. and Cattrysse, Dirk and others},
  title = {Ease of disassembly of products to support circular economy strategies},
  journal = {Resources, Conservation and Recycling},
  year = {2018},
  volume = {135},
  pages = {323--334},
  doi = {10.1016/j.resconrec.2017.06.022}
}

@article{al2024automated,
  title={Automated Disassembly of Battery Systems to Battery Modules},
  author={Al Assadi, Anwar and G{\"o}tza, Thomas and Gebhardta, Andreas and Mannu{\ss}a, Oliver and Meesea, Bernd and Wannera, Johannes and Singhaa, Soumya and Halta, Lorenz and Birkea, Peter and Sauera, Alexander},
  journal={Procedia CIRP},
  volume={122},
  pages={25--30},
  year={2024},
  publisher={Elsevier}
}

@Article{robotics11010018,
AUTHOR = {Han, Sangchul and Choi, Myoung-Su and Shin, Yong-Woo and Jang, Ga-Ram and Lee, Dong-Hyuk and Cho, Jungsan and Park, Jae-Han and Bae, Ji-Hun},
TITLE = {Screwdriving Gripper That Mimics Human Two-Handed Assembly Tasks},
JOURNAL = {Robotics},
VOLUME = {11},
YEAR = {2022},
NUMBER = {1},
ARTICLE-NUMBER = {18},
URL = {https://www.mdpi.com/2218-6581/11/1/18},
ISSN = {2218-6581},
DOI = {10.3390/robotics11010018}
}

@misc{peteck2024performance,
  author = {{Panasonic Eco Technology Center Co., Ltd.}},
  title  = {Business Performance (FY2024)},
  year   = {2024},
  url    = {https://panasonic.co.jp/peteck/performance.html},
  note   = {Accessed: 2026-02-14}
}

@inproceedings{Kiyokawa2026schedule,
author = {Takuya Kiyokawa and Tomoki Ishikura and Shingo Hamada and Genichiro Matsuda and Kensuke Harada},
title = {Hierarchical Planning and Scheduling for Reconfigurable Multi-Robot Disassembly Systems under Structural Constraints},
booktitle = {Proceedings of IEEE/SICE International Symposium on System Integration (SII)},
pages = {},
year = {2026}
}

@inproceedings{Xu2025EcoDesign,
  author    = {Xu, Junzhe and Mitake, Yuya and Matsumoto, Mitsutaka and Miyaji, Naoya and Hamada, Shingo and Matsuda, Genichiro and Tajima, Akio and Umeda, Yasushi},
  title     = {A Method for Deriving Remanufacturing Design Guidelines Based on Product Information and Lifecycle Scenarios},
  booktitle = {Proceedings of EcoDesign 2025 International Symposium},
  year      = {2025},
  month     = {November},
  number    = {A-10-02}
}

\end{document}